\documentclass{article}
\usepackage[nonatbib,final]{neurips_2023}
\usepackage[
    backend=biber,
    style=numeric-comp,
    sortcites=true,
    sorting=none,
    natbib=true,
    firstinits=true,
]{biblatex}
\usepackage{dblfloatfix}
\usepackage{caption}
\usepackage{subcaption}
\usepackage{listings}
\usepackage{wrapfig}
\usepackage{dramatist}
\usepackage{xspace}
\usepackage{pifont}
\usepackage{multirow}
\usepackage{tcolorbox}
\usepackage{longtable}
\usepackage{tikz}
\usepackage{booktabs}
\usepackage{datetime}
\usepackage{printlen}
\usepackage{lipsum}

\usepackage{amsfonts}
\usepackage{amsmath}
\usepackage{amssymb}
\usepackage{lineno}

\usepackage[bottom]{footmisc}

\usepackage{layouts}

\title{EarthPT: a time series foundation model\\for Earth Observation}

\author{%
    Michael J. Smith\thanks{mike.smith@aspiaspace.com} \quad Luke Fleming \quad James E. Geach\\
    Aspia Space Ltd., Cornwall, UK
}

\begin{document}

\maketitle

\begin{abstract}
    We introduce EarthPT -- an Earth Observation (EO) pretrained transformer.
    EarthPT is a 700 million parameter decoding transformer foundation model
    trained in an autoregressive self-supervised manner and developed
    specifically with EO use-cases in mind. We demonstrate that EarthPT is an
    effective forecaster that can accurately predict future pixel-level surface
    reflectances across the 400-2300\,nm range well into the future. For
    example, forecasts of the evolution of the Normalised Difference Vegetation
    Index (NDVI) have a typical error of approximately 0.05 (over a natural
    range of $-1\rightarrow1$) at the pixel level over a five month test set
    horizon, out-performing simple phase-folded models based on historical
    averaging. We also demonstrate that embeddings learnt by EarthPT hold
    semantically meaningful information and could be exploited for downstream
    tasks such as highly granular, dynamic land use classification.
    Excitingly, we note that the abundance of EO data provides us with -- in
    theory -- quadrillions of training tokens. Therefore, if we assume that
    EarthPT follows  neural scaling laws akin to those derived for Large
    Language Models (LLMs), there is currently no data-imposed limit to scaling
    EarthPT and other similar `Large Observation Models.' 
\end{abstract}

\section{Introduction}

Deep learning's current `hot topics' are foundation models in the vein of
EleutherAI's GPT-NeoX, OpenAI's GPT-{\it N} models, DeepMind's Chinchilla, and
the RWKV Foundation's eponymous model
\parencite{ref_gptneox,ref_brown2020gpt3,ref_openai2023gpt4,ref_hoffmann2022,ref_peng2023}.
These remarkably simple models contain a few standard deep learning building
blocks and are trained by repeatedly predicting the next item in a sequence.
Surprisingly, these models' performances scale with dataset and model size via
a simple power law \parencite{ref_cortes1993,ref_kaplan2020}. 
Even more astoundingly, at a certain scale of data and
compute, these models display `emergent abilities' such as apparent knowledge
of arithmetic, law, geography, and history \parencite[e.g.][]{ref_wei2022}.  In
March 2022 a team at Google DeepMind discovered that -- optimally -- the size
of these foundation models should be scaled in a roughly equal proportion to
the size of the dataset used to train them \parencite{ref_hoffmann2022}.
\textcite{ref_smith2023} demonstrated that this implies that the current
constraint on state-of-the-art textual foundation model performance is dataset
size, and not model size as previously thought. Although we are running out of
useful high quality textual data to train foundation models, there remains an
untapped abundance of high quality data in other domains
\parencite{ref_friel2022,ref_villalobos2022}. \textcite{ref_smith2023} argue
that astronomy is one such domain, and we argue here that remote sensing data
sets, and in particular Earth Observation (EO) spatial and temporal data, can
also be used as an additional non-textual data mode to aid in the training of
ever larger, more generalist, and more performant foundation models. 

Here we demonstrate that EO imaging data can be used to train a sizable
transformer model in the spirit of large language modelling. To this end we
train a Chinchilla-optimal 700M parameter decoding transformer model on 14B
tokens of EO data in the form of multispectral time series for just over one
hundred million individual pixels. The time series are analogous to word and
sentence sequences in textual models, but in this case represent surface-level
(solar) reflectance values measured in a number of passbands across the
400--2300\,nm spectral range -- i.e.\ the wavelengths corresponding to
traditional `optical' EO imagery. 

Single pixel time series are commonly used in remote sensing to train
transformer and self-attention based networks on supervised tasks
\parencite[e.g.][]{ref_garnot2020,ref_russwurm2020}.  However, 
currently few works apply these models in a self-supervised manner. Those
that do are typically limited to very short -- or even single
step -- time series inputs \parencite[e.g.][]{ref_cong2022,ref_reed2022}.
The closest approach in the literature to EarthPT is perhaps
\textcite{ref_tseng2023}. They show that an encoding transformer model
\parencite[i.e.][]{ref_bert} is capable of learning semantically meaningful
embeddings from remote sensing time series. Their model is trained on a
relatively small dataset comprised of 21.5M tokens arranged into chunks
of shape $\texttt{[time,channel]} \equiv \texttt{[12,19]}$.
\textcite{ref_tseng2023} note their model's capability despite its small size.
Our work has a diametrical and complementary purpose; we aim to demonstrate
that a transformer model trained on EO data is capable of scaling to the extent
that we have seen in the natural language domain, with similar potential for
wide utilisation and impact. In particular, we  demonstrate that EarthPT can
accurately forecast reflectance values well into the future, thus providing a
method to predict -- and therefore an opportunity to mitigate -- future events
associated with environmental threats such as drought.

\section{Methods} \label{sec_methods}

This section describes the datasets that we use to train EarthPT and the
hyperparameters and training routine of our chosen decoding transformer
architecture.

\paragraph{Training imagery.} \label{sec_sentinel}

ClearSky is a proprietary deep learning algorithm that accurately predicts the
equivalent of European Space Agency Sentinel-2 imagery products across 10
spectral bands: Blue, Green, Red, Red Edge 1-4, NIR, and SWIR 1 and 2. The
input data for ClearSky is Sentinel-1 C-band Synthetic Aperture Radar (SAR)
imagery at 10\,m/pixel resolution \parencite{ref_agram2022}. SAR imagery is
impervious to cloud cover, and the ClearSky algorithm allows us to construct
full multispectral imagery time series of Sentinel-2 equivalent reflectances
uninterrupted by cloud. In this work we generate ClearSky inferred imagery for
an area of interest in the UK defined by a $100\times100$\,km region
corresponding to the TL square of the British National Grid (BNG) reference
system. We define training and validation set time series range from January
2015 to December 2022, and test set time series ranges from January 2023 to May
2023. The time series are sampled at the same cadence as the observing pattern
of Sentinel-1, which for this location is five days on average.   

\paragraph{Preprocessing.} \label{sec_preprocessing}

We recompose the observation arrays into a set of \texttt{float16}
\texttt{NumPy} \parencite{ref_harris2020} arrays of shape
\texttt{[index,time,channel]}, where \texttt{index} corresponds to the
flattened spatial index of a $10\times10\,$km$^2$ BNG tile, \texttt{time}
corresponds to the date of  observation, and \texttt{channel} corresponds to
the individual spectral bands and the date embedding bands of the current and
next observation.  The date embedding is calculated via the equation
    $\hat{t} = \left(\sin\left(2\pi t/365\right), \cos\left(2\pi t/365\right)\right),$
where $t$ is the date of the observation in days since 1st January of the year
of observation.  The spectral band reflectances (originally on a 0--10,000
scale) are normalised as
    $\hat{v} = v/500 - 1,$
which keeps them approximately in the range $[-1,1]$.  We treat each temporal
observation as a separate `token', and therefore the TL training set (a subset
of the full UK data set) comprises approximately 100B tokens.  Once
constructed, we can efficiently access these data structures at train time via
memory-mapping. 

\paragraph{Transformer architecture.} \label{sec_transformer}

EarthPT is based on the autoregressive transformer architecture described in
\textcite{ref_radford2019gpt2}, with some alterations to accommodate our
non-textual dataset. In place of the usual word embedding routine we use a
multilayer perceptron to embed the input data so that it has the same
dimensionality as the time embedding vector. To provide the model with a
knowledge of the time of observation, we feed the network an additional pair of
float embeddings corresponding to the date of the current and next
observation. We train EarthPT in the usual autoregressive way, by repeatedly
predicting the next observation in a given set of time series. We train using
the Adam optimiser \parencite{ref_kingma2014}, and use the Huber loss
\parencite{ref_huber1964}. We trained a range of model sizes from 10M to 700M
trainable parameters, and we present the hyperparameters for all our models in
Appendix~\ref{sec_hyperparams}.  The remainder of this paper focuses on our
largest EarthPT model, EarthPT-700M.

In lieu of a domain-specific neural scaling law we use the Chinchilla neural
scaling law as a convenient rule-of-thumb to decide our dataset size.  This law
suggests that a compute-optimal decoding transformer model should be trained
roughly following the scaling
    $N \sim 20 D,$
where $N$ is the number of parameters in the model, and $D$ is the number of
tokens in the training set \parencite{ref_hoffmann2022}.  This corresponds to
14B tokens for our 700M parameter model.  To this end we train EarthPT-700M on
8 A100 PCIe 40GB GPUs for a cumulative total of 90,000 steps of 160,000 tokens
each, i.e.\ 560 A100-hours of computation time.

\section{Results} \label{sec_results}

We find that our EarthPT models share similar training behaviour with
traditional LLMs; further details of training runs can be
found in Appendix~\ref{sec_hyperparams}.  In this section we describe how
EarthPT-700M performs on the task of forecasting remote sensing data.

\begin{figure*}[h]
\centering
\includegraphics[width=0.9\textwidth]{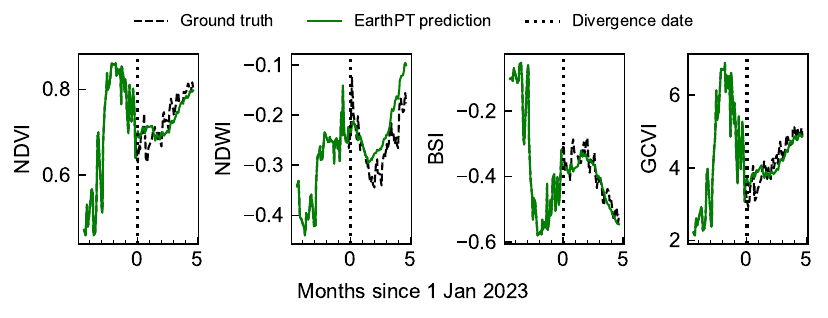}
\caption{Predictions of some common remote sensing indicators for a randomly
chosen pixel within the UK National Grid TL tile. We condition EarthPT on
ClearSky time series from 1st January 2015 to 1st January 2023, with outputs
after this divergence date constituting a long-term forecast to be compared to
the unseen observations.}
\label{fig_indices_}
\end{figure*}

\begin{wrapfigure}{r}{0.5\textwidth}
\vspace{-10pt}
\centering
\includegraphics[width=0.48\textwidth]{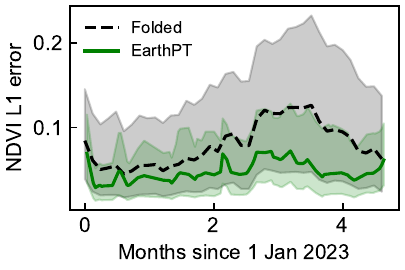}
\caption{Median L1 error and interquartile ranges of NDVI predictions for 1M
pixels in the TL63 tile. EarthPT long-term forecasts out-perform a simple
phase-folded model based on historical averages out to a horizon of five
months.}
\label{fig_indices_fanned}
\vspace{-10pt}
\end{wrapfigure}

Analogously to how autoregressive language models can be used to
generate text, we can use EarthPT to generate (i.e.\ forecast) future remote
sensing data, in this case the pixel-wise surface reflectance across the
optical-infrared. In Figure \ref{fig_indices_} we show EarthPT forecasts for
four representative remote sensing indices: Normalised Difference Vegetation
Index (NDVI), Normalised Difference Water Index (NDWI), Bare Soil Index (BSI),
and Green Chloropyll Vegetation Index (GCVI). These represent time streams of a
single pixel selected from the TL tile. Forecasting starts on the 1st of
January 2023 and runs to May 2023. We compare the forecast to the measured
values of these indices across this interval, which the model has not `seen'.
For brevity, we show a single pixel here, forecasting can be scaled across all
pixels to generate predicted imagery products.  

We can quantify performance by assessing the residual between the forecasted
and measured value of the parameter of interest (e.g.\ NDVI) as a function of
look-ahead time. Figure~\ref{fig_indices_fanned} shows the median L1 error for
$\sim$$10^6$ pixels in BNG tile TL63, up to five months into the future. This
is compared to a prediction based on a phase-folded model which comprises an
average annual time series constructed from 7 years of historical data. We find
that EarthPT has a median L1 error across all time of 0.05 and the folded model
has a median L1 error of 0.08, noting that NDVI has a natural range of
$-1\rightarrow 1$. We can conclude that EarthPT out-performs a phase-folded
model consistently over the forecast window, delivering actionable predictions
on key remote sensing indices (such as NDVI) that could be used, for example,
in the prediction of drought conditions well in advance
\parencite{ref_salakpi2022}.


\section{Future Work} \label{sec_future}

\begin{figure}[htb]
\centering
\includegraphics{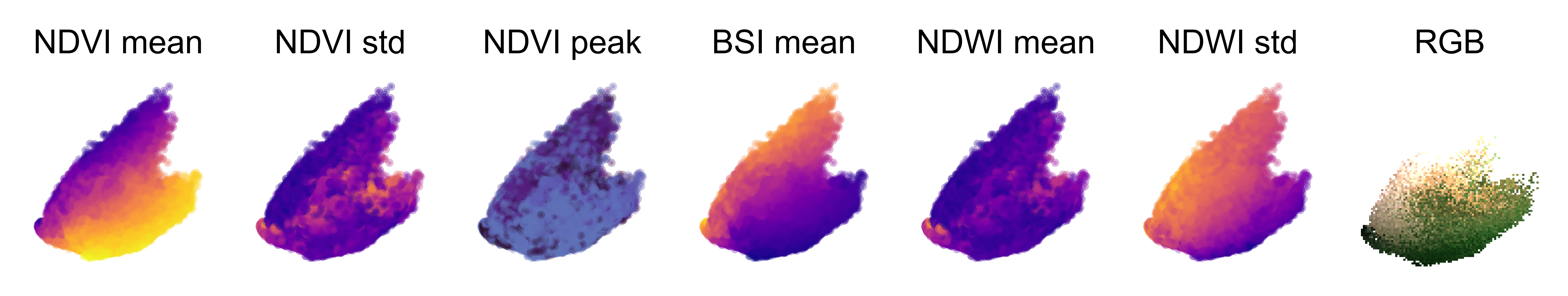}
\caption{
    EarthPT embeddings for the two million pixel time series located on the
    TL63 and TL64 BNG tiles. We colour each scatter plot with a different set
    of emergent remote sensing index values. `RGB' is the colour of a pixel in
    that part of the embedding space at the height of the summer of 2022.
    `Mean' is the mean of a given index across the 2022 calendar year, and
    `std' is the standard deviation of the index across the year. `NDVI peak'
    is the time of the year corresponding to maximum NDVI; darker values are in
    the winter, and lighter values are in the summer. Note the coherent
    structure in the projected embedding space.
    }
\label{fig_embeddings}
\end{figure}

Foundation models are notoriously flexible, and so one can envision myriad
downstream tasks. In the field of geospatial data analysis, we can consider how
EarthPT could be deployed for land cover classification. To illustrate, we
generate representation embeddings by extracting the outputs of the penultimate
neuronal layer and obtain the embedding of a pixel's time series by simply
taking the mean of all of its output embeddings (one embedding is output at
each time step). Each embedding has a dimensionality of 1280, but we can
visualise them by projecting onto a two-dimensional manifold. We use principle
component analysis (PCA) as our projection technique
\parencite{ref_pearson1901}.  Figure~\ref{fig_embeddings} shows a selection of
emergent remote sensing indices (introduced above) for a set of embeddings of
time series across 2022. By colour-coding the projected embedding space we see
that it has a physically meaningful organisation, with coherent structure of,
for example, the time-averaged NDVI, BSI, RGB, etc. If we were to cluster and
calibrate the embedding space with semantically meaningful labels (e.g.\ crop
type, growth stage, event) this could be used to create a dynamic and highly
granular land cover classifier. Furthermore, we anticipate that fine-tuning
with EarthPT-learnt embeddings will be beneficial for a range of downstream
tasks \parencite[see for example][]{ref_graff2014,ref_bert}. One could imagine
training EarthPT to produce a single embedding space for all EO (and other)
multi-modal data types \parencite{ref_reed2022gato}. This would be remarkably
powerful tool for interpreting remote sensing data, where we foresee diverse
applications in a range of sectors, from agriculture to insurance and beyond.

While useful as a rule-of-thumb, the Chinchilla scaling laws may not be
suitable for EO datasets, and so follow-up work will derive a specific scaling
law for our ClearSky dataset. This in turn will give us a solid theoretical
grounding for further scaling of EarthPT, allowing us to train a significantly
larger model. For example, with our ClearSky model for the UK we have access to
4.3T (trillion) tokens that could be used to train EarthPT, and when
considering larger geographic coverage we theoretically have access to over a
quadrillion tokens. Compute cost aside, we could safely train a 50T parameter
model on this data, assuming that our model scaling roughly follows the
Chinchilla scaling law. This 50T parameter model would be around three orders
of magnitude larger than the current largest optimally-trained models
\parencite{ref_hoffmann2022,ref_touvron2023}.  Consequently, unlike traditional
LLMs, EarthPT and other similar `Large Observation Models' are far from their
theoretical data limit \parencite{ref_smith2023,ref_leung2023}.

\section{Conclusions} \label{sec_conclusions}

Inspired by the recent explosion of interest in LLMs, we present an Earth
Observation foundation model trained on time series taken from our ClearSky
generative algorithm. Our EarthPT Large Observation Model is capable of
forecasting surface level optical reflectance (and therefore a wide range of
common remote sensing indices) at the pixel level, months into the future.
EarthPT can also produce semantically meaningful embeddings for an input time
series, and we show that these capture useful information that could be
exploited for land cover classification, amongst other downstream tasks. We are
developing these applications and improving and extending EarthPT as part of
ongoing R\&D.  Excitingly, the number of tokens available for training is of
order $10^{15}$, so we are not currently data constrained. If neural scaling
laws hold, then improving EarthPT (and similar Large Observation Models) is a
solved problem: it is a simple matter of scaling data and compute.

\section*{Data and code availability}

The model weights and code are available on Zenodo at
\url{https://doi.org/10.5281/zenodo.10489724} and on Github at
\url{https://github.com/aspiaspace/EarthPT}. Please contact Aspia Space
directly at \url{contact@aspiaspace.com} for access to the ClearSky dataset.

\section*{Acknowledgements}

This project is part-funded by the UK Government through the UK Shared
Prosperity Fund. Cornwall Council has been chosen by Government as a Lead
Authority for the fund and is responsible for monitoring the progress of
projects funded through the UK Shared Prosperity Fund in Cornwall and the Isles
of Scilly.

\vspace{1em}

\includegraphics[height=22px]{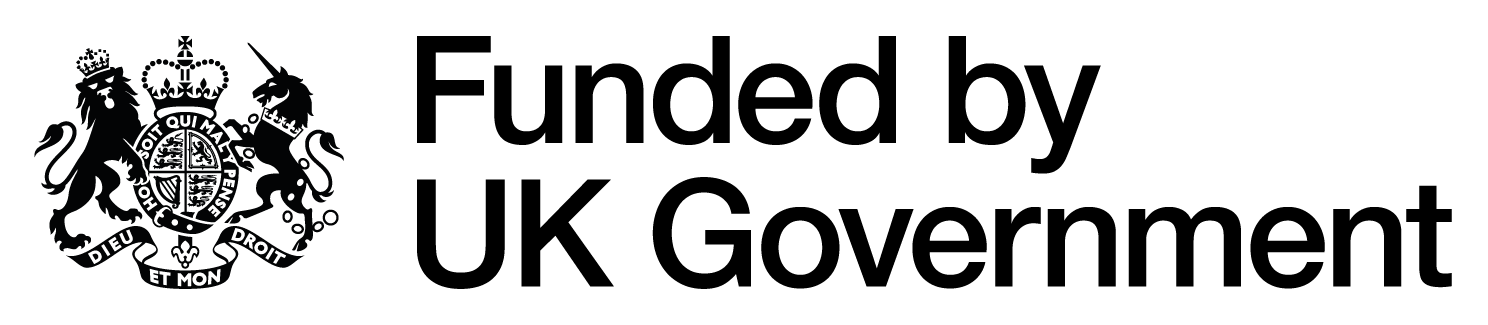}\hfill
\includegraphics[height=22px]{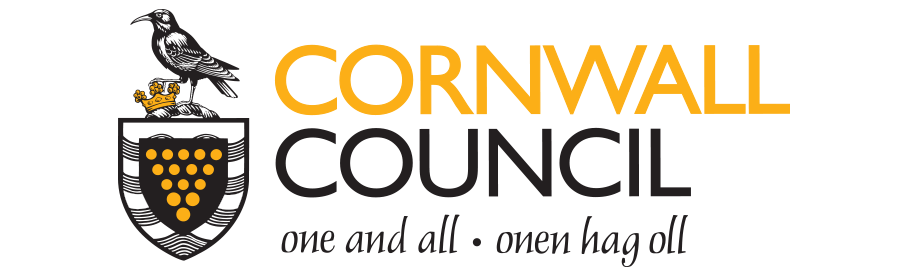}\hfill
\includegraphics[height=22px]{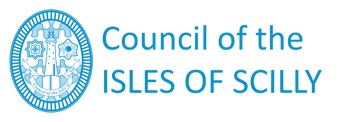}\hfill
\includegraphics[height=22px]{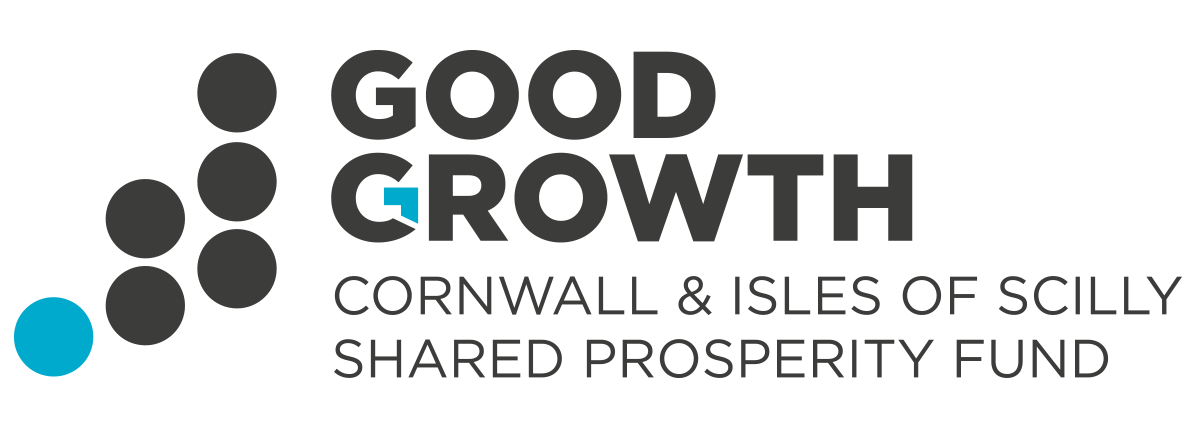}\hfill
\includegraphics[height=20px]{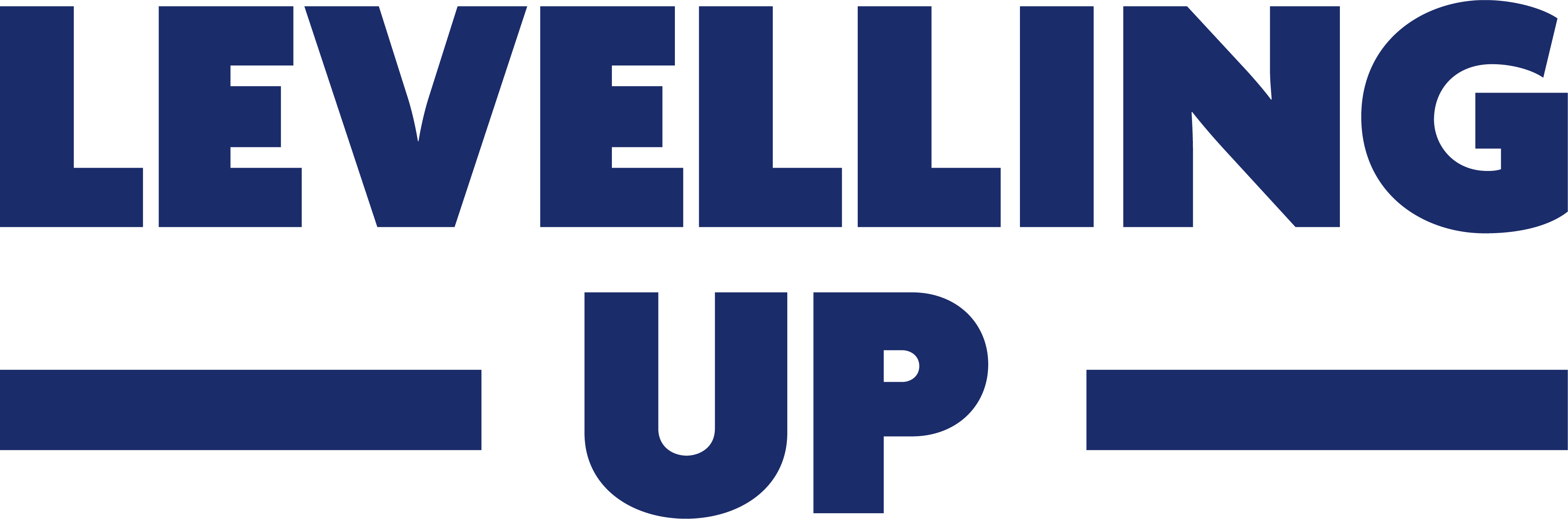}

\printbibliography[]

\vfill
\pagebreak

\appendix

\section{Carbon emissions}

The training of deep learning models requires considerable energy, contributing
to carbon emissions.  We trained EarthPT-700M on 8 A100 PCIe 40GB GPUs.  A
cumulative total of 560 A100-hours of computation was performed, corresponding
to a GPU energy expenditure of $145\,$kWh.  Assuming a carbon efficiency of
$0.193\,\text{kg}\,\text{CO}_2\text{eq.}/\text{kWh}$ for the UK, we estimate
our emissions as ${\sim}30\,\text{kg}\,\text{CO}_2\text{eq.}$ We conducted
these estimations via the excellent machine learning impact calculator
presented in \textcite{ref_lacoste2019}.

\section{Hyperparameters and scaling tests} \label{sec_hyperparams}

\begin{figure}[htbp]
\centering
\includegraphics{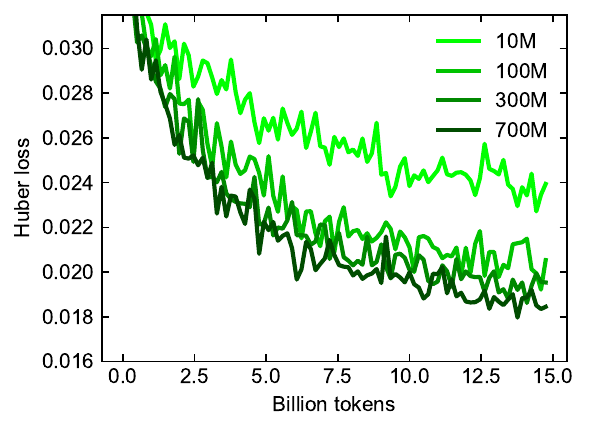}
\caption{Loss curves for our various EarthPT training runs.}
\label{fig_loss}
\end{figure}

Hyperparameters used to train all our EarthPT models are shown in
Table~\ref{tab_hyperparam}, and our training run loss curves for all our model
sizes are shown in Figure~\ref{fig_loss}.  We can see in Figure~\ref{fig_loss}
that the larger models are still learning at the end of training, and so we
expect that training larger models on more data would improve performance.

\begin{table}[htbp]
\centering
\caption{Hyperparameters used to train our EarthPT models. Following
\textcite{ref_hoffmann2022}, we decay the learning rate by a factor of 10 over
a horizon of a length $1.1\times$ the total training steps.\\}
\label{tab_hyperparam}
\begin{tabular}{l|rrrr}
    \toprule
    & \multicolumn{4}{c}{EarthPT model size} \\
    Hyperparameter & 700M & 300M & 100M & 10M \\
    \midrule
    Number of layers & 36 & 26 & 20 & 10 \\
    Number of heads  & 20 & 16 & 10 & 10 \\
    Embedding dimension & 1280 & 1024 & 640 & 320 \\
    Block size & 256 & 256 & 256 & 256 \\
    Batch size & 0.164M & 0.164M & 0.164M & 0.164M \\
    Total training steps & 90,000 & 90,000 & 90,000 & 90,000 \\
    Max learning rate & 2E-5 & 2E-5 & 2E-5 & 2E-5 \\
    \bottomrule
\end{tabular}
\end{table}

\end{document}